# Using Biological Variables and Social Determinants to Predict Malaria and Anemia among Children in Senegal

Boubacar Sow[†]　　　Hiroki Suguri[††]　　　Hamid Mukhtar[†††]　　　Hafiz Farooq Ahmad[††††]

[†] Graduate School of Project Design, Miyagi University, 1-1 Gakuen, Taiwa Cho, Kurokawa Gun, Miyagi Prefecture, 981-3298 Japan

[††] School of Project Design, Miyagi University, 1-1 Gakuen, Taiwa Cho, Kurokawa Gun, Miyagi Prefecture, 981-3298 Japan

[†††] School of Electrical Engineering and Computer Science, National University of Sciences and Technology, NUST Sector H-12, Islamabad, 44000 Pakistan

[††††] Department of Computer Science, College of Computer Sciences & Information Technology, King Faisal University, P. O. Box: 400 Al-Hassa, 31982 Kingdom of Saudi Arabia

E-mail: [†] p1752007@myu.ac.jp, [††] suguri@myu.ac.jp, [†††] hamid.mukhtar@seecs.edu.pk, [††††] farooq.ahmad@seecs.edu.pk

**Abstract:**　Integrating machine learning techniques in healthcare becomes very common nowadays, and it contributes positively to improving clinical care and health decisions planning. Anemia and malaria are two life-threating diseases in Africa which affect the red blood cells and reduce hemoglobin production. This paper focuses on analyzing child health data in Senegal using four machine learning algorithms in Python: KNN, Random Forests, SVM and Naïve Bayes. Our task aims to investigate large scale data from The Demographic and Health Survey (DHS) and to find out hidden information for anemia and malaria. We present two classification models for the two blood disorders using biological variables and social determinants. The findings of this research will contribute to improving child healthcare in Senegal by eradicating anemia and malaria, and decreasing child mortality rate.

## 1. Introduction

Malaria and anemia are disorders of the red blood cells common in African countries. The first one is caused by Plasmodium parasites from the bite of infected female anopheles mosquitoes. The second one is caused by an iron deficiency to make enough hemoglobin (protein molecule in the red blood cell). Malaria and anemia can be classified using independent variables such as morphometric parameters of the blood (abnormal condition of the red blood cells morphology), environmental factors, living condition, vaccination history, characteristics (height, weight, sex) etc. In many of the cases, a simple test of thick and thin blood smears can determine whether you have anemia or malaria.

Strong relationship exists between malaria and anemia because both affect the red blood cells. If malaria becomes severe, the number of red blood cells decreases and it can cause a severe anemia. In such a case, it is known as malarial anemia. We hypothesized that using two datasets from the same data source and having some common variables are innovative approach to classify the two diseases.

The challenge of this paper is to deal with biological and social variables to classify the two blood diseases and lay out possible solutions for malaria and anemia eradication. According to WHO's study [1], two actions must be taken on social determinants of health: improving daily living conditions and tackling the inequitable distribution of power, money and resources. Therefore, one possible hypothesis is that by improving life style condition and children's nutritional habits, it is possible to fight against life-threating diseases in African countries including Senegal. Indeed, this paper is an added value of understanding the problem, and assess probable actions to take in child health sector.

### 1.1. Problem Statement

Today, supervised learning claims to be a good way to improve healthcare services. One of the most well-known techniques is disease classification, that uses various kinds of independent variables from patient diagnosis data or electronic medical record databases. However, in the case of seasonal diseases such as malaria and nutritional deficiencies such as anemia, the use of social determinants and biological variables should be an efficient option for disease prediction. This is because of two reasons. In one hand, poor quality of life is one of the major causes of health disabilities. In another hand, biological variables are considered as critical factors affecting child health.

Therefore, the classification outcomes using those types of variables will allow policy makers to solve health inequity problems and plan new strategies to control and eliminate malaria and anemia.

## 1.2. Purpose of the Research

The primary concern of this paper is solving health perspective problems and seeking knowledge for better health. It aims to investigate a large-scale data and show the importance of using social and biological variables to predict malaria and anemia.

## 2. Methods
## 2.1. Data Collection

Data has been collected from the Demographic Health Survey (DHS), which was made from 2015 to 2016 in Senegal. The initial dataset contains 986 variables and 6935 instances. Attributes include household, child birth record, medical history, immunization record, etc. The survey is one of the household survey series of the DHS. It was made during two seasons: September to January and February to August. The first season is the end of the rainy season in Senegal, which is a hot and humid period. The second season is known as the dry season with a dry north-eastern wind.

The survey was conducted using three types of questionnaires: a household questionnaire and two individual questionnaires for women (15 to 49 years) and for men (15 to 59 years). Information collected is related to household characteristics and members, anthropometric measurements, socio-demographic characteristics, pregnancy and postnatal care, woman's occupation and status, immunization and nutrition. In addition, a blood sample was taken from children under 5 years to carry out a hemoglobin level measurement and to estimate the prevalence of malaria.

## 2.2. Data Pre-processing

Our initial dataset from DHS contained a large number of unimportant variables for disease prediction. Most of irrelevant variables for our study are somehow related to respondent's identifiers, husband's information, cluster sampling information and redundant variables. We prepared correctly two datasets by removing the unnecessary data and choosing relevant variables and their appropriate representations. This step known as data pre-processing includes data reduction and data transformation.

### 2.2.1 Feature Engineering

The concept of feature engineering is a step on data mining to select relevant attributes for predictive modelling and to improve the model accuracy. For this purpose, we started by removing variables which have many missing values. Secondly, we applied three feature engineering techniques.

- Correlation: This technique includes removing attributes highly correlated to avoid redundancy or similarity. In such case variables with a strong correlation greater than 0.75 are removed in the dataset.

- Recursive Feature Elimination (RFE): The above technique has been applied and we chose the best performing feature and then repeated the same process recursively with the remaining attributes until all attributes are ranked.
- Principal Component Analysis (PCA): The purpose of this technique is to reduce the dimensionality of our data. We computed all principal components and selected those which explained the maximum variance. The idea behind this technique is to illustrate for each principal component how it explains the variability of the data and the correlation with the corresponding variable.

The use of these variable selection techniques is justified because of its simplicity, scalability and good empirical success [2]. The main purpose of using correlation criteria and PCA is that they easily detect the dependencies between two variables or between input and output variables. Therefore, the data reduction task is done by removing the highly correlated or the less explained variance. RFE displays a variable ranking to find which variable is more relevant than other.

Table 1 describes the data reduction process and the number of variables removed in each step. The final dataset for anemia contains 2583 instances and 20 attributes, and for malaria 62 instances and 17 attributes.

| Techniques | Number of variables removed |
|---|---|
| Missing values and unimportant variables | 940 |
| Correlation | Anemia dataset (11), Malaria dataset (17) |
| RFE | Anemia dataset (10), Malaria dataset (5) |
| PCA | Anemia dataset (2), Malaria dataset (7) |

Table 1: Number of variables removed in each step

2.2.2. **Data Transformation**

This step involves representing the data in numerical format and setting up a label that is understandable and interpretable by our models. As we use the same data source for malaria and anemia, Table 2 shows common variables used for both classifications.

| Type | Attributes | Representation |
|---|---|---|
| **Biological Variables** | Sex of child | Male - Female |
| | Child height | Numeric |
| | Currently amenorrheic | No - Yes |
| | Number of antenatal visits during pregnancy | Numeric |
| **Social Determinants** | Region | Dakar, Ziguinchor, Diourbel, Matam Saint-Louis, Tambacounda, Thies, Kaolack, Louga, Fatick, Kolda, Kaffrine, Kedougou, Sedhiou |
| | Type place of residence | Urban - Rural |
| | Highest educational level | No Education - Primary Secondary - Higher |
| | Source of drinking water | Piped into dwelling, Piped to yard, Public tap, Tube well, Protected well, unprotected well, River or lake, Cart with small tank, Bottled water |
| | Respondent's occupation | Not working, Sales, Agricultural, self-employed Household and domestic, Services, skilled manual, Unskilled manual |

Table 2: Common variables anemia and malaria

**a. Setting Label Attribute for Anemia**

The process of selecting a class label is an important step in supervised learning. It is an attribute to predict based on other independent variables or predictors. In this research, the model is to predict anemia level using important predictors related to children's biological variables and social determinants. Table 3 shows anemia status definition according to the World Health Organization:

| Level | Definition |
|---|---|
| Severe | Hemoglobin level less than 7g/dl |
| Moderate | Hemoglobin level between 7 and 9.9 g/dl |
| Mild | Hemoglobin level between 10 and 10.9 g/dl |
| Not Anemic | Hemoglobin level greater or equal to 11 g/dl |

Table 3: anemia status definition

As we are focusing on disease classification, it becomes thoughtful to study the normal and abnormal condition or the presence and absence of a disease. Therefore, it is a binary classification task. To set a binary class, a final transformation must be made by grouping Severe, Moderate and Mild as POSITIVE and Not anaemic as NEGATIVE.

In addition to variables in Table 2, Table 3 shows the supplement variables used for anemia classification.

| Type | Attributes | Representation |
|---|---|---|
| **Biological Variables** | Received Measles | No, Vaccination date on card, Reported by mother, Marked on card, Don't know |
| | VitaminA in last 6 months | No, Yes, Don't Know |
| | Currently breastfeeding | No - Yes |
| | Drugs for intestinal parasites in last 6 months | No, Yes, Don't Know |
| | Anaemia level | **Positive** ->Severe, Moderate, Mild **Negative** ->Not anaemic |
| **Social Determinants** | Drank from bottle with nipple last night | No – Yes – Don't Know |
| | Number of times ate solid, semi solid or soft food yesterday | None – 7+ - Don't know |
| | First 3 days given infant formula | No – Yes |
| | First 3 days given tea, infusions | No – Yes |
| | First 3 days given other | No – Yes |
| | Has health card | No Card, Yes seen, Yes not seen, No longer has card |

Table 4: Supplement anemia dataset

**b. Setting Label Attribute for Malaria**

The same process of feature engineering is applied in dataset to select relevant attributes to predict malaria result test. During the survey, malaria measurement was done by testing the thin blood smear. So, we set this attribute "result of malaria test" as a label for predicting malaria. Table 5 shows the supplement variables used for malaria classification.

| Type | Attributes | Representation |
|---|---|---|
| Biological Variables | Respondent's current age | Numeric |
| | Currently pregnant | No - Yes |
| | Received Vitamin A dose in first 2 months | No – Yes – Don't Know |
| | Had fever in last two weeks | No – Yes |
| | Had cough in last two weeks | No – Yes |
| | Result of Malaria test | Positive Negative |
| Social Determinants | Season of Interview | Rainy season (Sep to January) Dry season (February to August) |
| | Household has Electricity | No - Yes |

Table 5: Supplement malaria dataset

## 2.3. Building Model for Malaria and Anemia

Predictive analytics becomes a new trend in health informatics. Several machine learning techniques and statistical tools can be used to build a prediction model. Most of them are already deployed in Python or R.

This research deals with four machine learning algorithms because each of them has many properties that make our binary classification accurate.
- K Nearest Neighbors (KNN)
- Support Vector Machine (SVM)
- Random Forest
- Naïve Bayes

Of course, several machine learning techniques can be applied in our dataset. Therefore, based on Ockham's Razor strategies [3], we choose those four algorithms because of their simplicity and capability to fit our data.

### 2.3.1. KNN

KNN is a powerful algorithm widely used in healthcare domain to predict heart diseases or breast cancer. It is a classifier that decides on the class of an unlabeled sample based on its k nearest labelled neighboring samples according to some distance measure such as Euclidian distance or Manhattan distance. The more the value of k is adjusted, the more the accuracy is getting greater and the overfitting problem goes away [4]. In the case of binary classification problem, let us assume $x_t$ be our target value.

The purpose of KNN algorithm is to find the k examples that are nearest to $x_t$ and k is always chosen to be an odd number.

### 2.3.2. SVM

The support vector machine or SVM framework is currently the most popular approach for "off-the-shelf" supervised learning [5]. It is based on kernel principle and separates input data linearly into two classes. SVM is typically an ideal machine learning algorithm for binary classification task.

Conventionally we reproduce the traditional notation used to compute SVM: +1 for positive label and -1 for negative label. Let us consider an input space $X = \{x_1, x_2, ..., x_n\}$ and an output label $Y = \{y_1, y_2\}$ where $y_1 = +1$ and $y_2 = -1$. Then the separator hyperplane is defined by:

$\{x : \mathbf{w}.\mathbf{x} + b = 0\}$ where w is the weight vector and b is a parameter or bias.

To solve this problem, we can easily verify the sign of $\{\mathbf{w}.\mathbf{x} + b\}$. If $\mathbf{w}.\mathbf{x} + b \geq 0$ then we assign an instance to the positive label (y = +1). Otherwise, the negative label (y = -1) is assigned (Figure 1). The distance between the two hyperplanes is known as the margin given by: $\frac{2}{||w||}$.

The purpose of the SVM is to compute the optimum margin to correctly separate the data into two classes. To do so, we face to an optimization problem, which is to maximize $\frac{2}{||w||}$ or minimize $||w||^2$. [8]

Max $\frac{2}{||w||}$ subject to $\mathbf{w}.\mathbf{x}_i + b \begin{cases} \geq +1 \text{ if } y_1 \\ \leq -1 \text{ if } y_2 \end{cases}$ for i=1, ..., n

In the case of non-separable data, we introduce a positive variable ξ called "slack" variable to correct the linearity problem. Then the optimization task becomes:

**min** $||w||^2 + C \sum_i^n \xi i$ (where C is the a regularization parameter) subject to : $y_i(\mathbf{w}.\mathbf{x}_i + b) \geq 1 - \xi i$ for i=1,..., n.

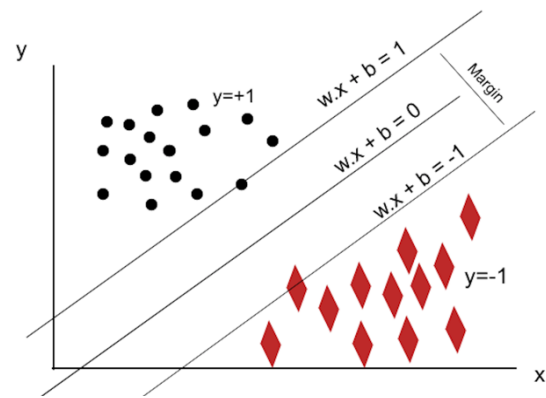

Figure 1: SVM

### 2.3.3. Random Forests

Random Forests is well known in healthcare area for disease prediction and outputs generally a good accuracy in binary classification. It is basically an ensemble of independent decision trees and it generates a bunch of random decision trees automatically.

In a binary classification task, we assume input variables X and output classes Y such that Y = {$y_1$, $y_2$} representing the different classes. The purpose is to model the posterior probability distribution P(**X**|**Y**), where **X** belongs to X and **Y** belongs to Y. The label predicted is given by computing the maximum a posteriori [6]:

$$Y_{max} = argmax_{Y \in Y}$$

### 2.3.4. Naïve Bayes

Naïve Bayesian classifier is one of the most efficient algorithms applied in disease prediction and has been used in artificial intelligence and medical diagnosis since the 1960s. [3] It is based on Bayes' rule which computes the conditional probability **P**(cause|effect) and quantifies the relationship between cause and effect.

Simply let us consider an ensemble of variables **X** be predictors represented by ($x_1$, $x_2$, …, $x_n$), and **Y** ensemble of class label where **Y**={$y_1$, $y_2$} representing the positive and negative values for binary classification. The probability of E being a class C is computed by [7]:

$P(Y \mid X) = \frac{P(X|Y)P(Y)}{P(X)}$

i.e   $P(Y|x_1, x_2, …, x_n) = \frac{P(x_1,x_2,…,x_n|Y)P(Y)}{P(x_1,x_2,…,x_n)}$

Therefore, the positive class probability is given by: $P(Y=+ \mid x_1, x_2, …, x_n)$

and the negative class by: $P(Y=- \mid x_1, x_2, …, x_n)$.

### 3. Results

Tables 5 and 6 show high accuracy when running SVM to classify both malaria and anemia. SVM classifies more consistently than others the decision boundaries between the positive and negative test result of malaria, in the same way, the binary level for anemia. In simple words, SVM is a good classifier for a two-dimensional outputs label (positive and negative).

Let us analyze the impact of our independent variables when running SVM. Most of the input variables for malaria are related to the status of child's mother, vaccination and illness histories, the season of interview and the place of residence. And for anemia, variables are related to first 3 days' nutritional habits, vaccination history, the place of residence and the quality of drinking water. They are non-morphometric parameters, but using linear SVM (with Cost=10), they predict accurately both blood disorders.

In summary, it is achievable to classify the abnormality of red blood cells by using social determinants such as the household information, child vaccination history, nutritional factors and biological variables such as height, age and sex.

| Algorithms | Accuracy |
|---|---|
| KNN | 69.23% |
| Random Forest | 76.92% |
| **SVM** | **92.03%** |
| Naïve Bayes | 84.61% |

Table 5: Algorithms comparison for Malaria

| Algorithms | Accuracy |
|---|---|
| KNN | 79.11% |
| Random Forest | 82.83% |
| **SVM** | **93.41%** |
| Naïve Bayes | 76.64% |

Table 6: Algorithms comparison for Anaemia

### 4. Discussions

Binary classification is widely used in medical diagnosis or medical testing to determine the presence or absence of disease because of its capability to separate the data into two groups. The findings of this research suggest the use of Support Vector Machine as a binary classifier for both malaria and anemia. The logic of SVM is to divide the hyperplane into two subsets of Positive and Negative. This method presents accurate result in medical prediction. It was being applied previously in heart disease or breast cancer prediction [9]. Using SVM for malaria or anemia prediction helps us to minimize the empirical error and improve the classification accuracy. The biggest problem of SVM in binary classification is the choice of the kernel and the cost parameter. Using scikit-learn package for python, however, this challenge is largely achieved because of the optionality of kernel. If the user does not specify a kernel, a default kernel (radial basis function kernel) will be chosen.

The choice of biological variables and social determinants in disease prediction or risk estimation is quite normal because, in all countries, human health condition depends on where people live, how they work or what they eat. In both datasets used in this study, all variables show high importance of building our prediction model. The findings of this research claim a simple way to predict malaria and anemia without red blood cells measurement. This current study is an asset in reducing child mortality rate and reorient physicians to more effective way to prevent malaria propagation or anemia.

In this study, some variables are related to mother's current conditions such as currently pregnant or currently amenorrhea. They play an important role when it comes to analyze child health situation. They may be a good indicator for malaria and anemia measurements among children.

Furthermore, combining biological variables and social determinants should be applicable in both rural and urban area when collecting the data. This feature in the data collection step helps researchers to gain a heterogeneous sample of data.

This study is a knowledge discovery to tackle child health problems, particularly in terms of resources distribution and living environment. The biggest limitation of this study is the small sample size of malaria dataset owing to a big number of missing values.

## 5. Conclusion and Future Work

Anemia and malaria are two public health problems for two vulnerable populations: pregnant women and children under 5 years. Then, it is substantial to figure out the impact of biological variables and social determinants for such kind of diseases. This study presented a data classification problem solving for better child health, which can be used in a strategic planning level to solve health inequities. We performed four machine learning techniques for binary classification in two different datasets and compared the outputs. The results showed high accuracy when using SVM for both classification problems.

We believe that the current research is an innovative application of machine learning techniques as we used social and biological factors to predict accurately malaria and anemia.

In perspective, we plan to develop a new model to classify the severe stage of malaria known as malarial anemia because it is this stage that can cause a severe damage including death for children.

This paper did not address the multiclass classification problem and the possibility to use the same kind of variables for other disease predictions. Then, this specific problem area is still open for future work in solving disease classification problems by using social and biological factors.

## 6. Acknowledgments

The authors thank Japan International Cooperation Agency for its Master's Degree and Internship Program of African Business Education Initiative for Youth (ABE Initiative) to support the research and related activities.